\documentclass[lettersize,journal]{IEEEtran}
\usepackage{amsmath,amsfonts}
\usepackage[ruled,linesnumbered]{algorithm2e}
\usepackage{graphicx,color}
\usepackage{textcomp}
\usepackage{listings}
\usepackage{xcolor}
\usepackage{hyperref}
\usepackage{multirow}
\usepackage{booktabs}
\hypersetup{hidelinks=true}
\usepackage{orcidlink}
\usepackage{titlesec}
\usepackage{array}
\usepackage{textcomp}
\usepackage{stfloats}
\usepackage{url}
\usepackage{verbatim}
\usepackage{graphicx}
\usepackage{cite}
\usepackage[caption=false,font=normalsize,labelfont=sf,textfont=sf]{subfig}
\titlespacing*{\section}{0pt}{0.8em}{0.5em}
\titlespacing*{\subsection}{0pt}{0.6em}{0.4em}
\titlespacing*{\subsubsection}{0pt}{0.3em}{0.2em}

\begin{document}

\title{Towards Reliable Aerial–Ground Vehicle Collaboration: An Integrated Planning and Autonomy Framework for Field Deployment}

\author{Md Safwan Mondal$^{\orcidlink{0000-0003-4757-9403}}$, Luca Russo$^{\orcidlink{0009-0009-3450-4534}}$,\\ James D. Humann$^{\orcidlink{0000-0003-3858-3873}}$, James M. Dotterweich, and Pranav Bhounsule$^{\orcidlink{0000-0002-7504-6009}}$
\thanks{$^{1}$Md Safwan Mondal (corresponding author), Luca Russo and 
Pranav A. Bhounsule are with the Department of Mechanical
and Industrial Engineering, University of Illinois Chicago, IL,
60607 USA. {\tt\small mmonda4@uic.edu}, {\tt\small lrusso5@uic.edu }, {\tt\small pranav@uic.edu} $^2$James D. Humann is with DEVCOM Army Research Laboratory, Minneapolis, Minnesota, USA.{\tt\small james.d.humann.civ@army.mil}
   $^3$ James M. Dotterweich, is with DEVCOM Army Research Laboratory, Aberdeen Proving Grounds, Aberdeen, MD 21005 USA. {\tt\small james.m.dotterweich.civ@army.mil}}%
   \thanks{ *This work was supported by ARO contract number W911NF-24-2-0018. }
}

\markboth{Journal of \LaTeX\ Class Files,~Vol.~14, No.~8, August~2021}%
{Shell \MakeLowercase{\textit{et al.}}: A Sample Article Using IEEEtran.cls for IEEE Journals}


\maketitle

\begin{abstract}

Limited flight endurance significantly restricts the operational range of unmanned aerial vehicles (UAVs) in long-duration missions such as surveillance and inspection, where multiple spatially distributed Areas of Interest (AOIs) must be visited. These tasks require efficient routing—determining the sequence of visits—which directly impacts mission time, energy consumption, and overall feasibility. Pairing UAVs with unmanned ground vehicles (UGVs) for mobile recharging offers a promising solution, but introduces a tightly coupled cooperative routing problem involving UAV route planning, UGV road-constrained movement, energy management, and rendezvous scheduling under uncertainty. In this work, we present an integrated planning and autonomy framework for reliable field deployment. We formulate the problem as an energy-constrained cooperative routing task and solve it using a Deep Reinforcement Learning (DRL)-based planner that jointly optimizes the UAV visitation sequence and rendezvous locations with the UGV, outperforming baseline heuristics in minimizing total mission time. To bridge the gap between planning and execution, we introduce a standardized two-layer YAML-based mission API that captures environment states and structures lightweight, synchronized action sequences. This framework is supported by a complete autonomy stack using PX4/MAVSDK for UAV control and ROS 2/Nav2 for UGV navigation. Furthermore, we propose a lightweight Rendezvous-Aware Replanner (RARP) that operates online to handle environmental uncertainties, reducing energy margin violations from 83.33\% to 20.00\%. The full system is validated through outdoor field experiments, demonstrating robust cooperative navigation and adaptability in dynamic tasks, including a search-and-rescue scenario with vision-language model (VLM)-based hazard detection.

\end{abstract}


\section{INTRODUCTION}

\IEEEPARstart{T}he collaboration between Unmanned Aerial Vehicles (UAVs) and Unmanned Ground Vehicles (UGVs)
has been widely explored across domains that benefit from the complementary strengths of
aerial mobility and ground-level endurance and payload capacity. UAVs provide rapid coverage
and high-resolution sensing but are severely constrained by limited flight endurance. UGVs, in
contrast, offer long-duration operation, higher payload capability, and reliable navigation
along road networks. When integrated, a UGV can act as a mobile recharging or support platform, allowing the UAV to operate for longer durations by periodically  recharging from the UGV. This cross-domain synergy has motivated a growing body of research in fields such as search and rescue (SAR), precision agriculture, logistics, and infrastructure inspection.

In SAR operations, UAVs rapidly map disaster zones and
identify victims, while UGVs leverage aerial information for safe path planning, material
delivery, and ground-level intervention ~\cite{mehmood2018multi, john2025semi}. Beyond emergency response, UAV–UGV teams have been
applied to precision agriculture for large-scale monitoring
and data acquisition ~\cite{wei2019air, das2015devices}; to logistics for improving last-mile delivery
efficiency~\cite{wang2017vehicle, murray2015flying}; and to civil and
environmental domains for power-line inspection, infrastructure monitoring, and ecological
surveillance~\cite{hu2023use, hament2018unmanned}.

\begin{figure*}[]
\centering
\includegraphics[width=1.0\textwidth]{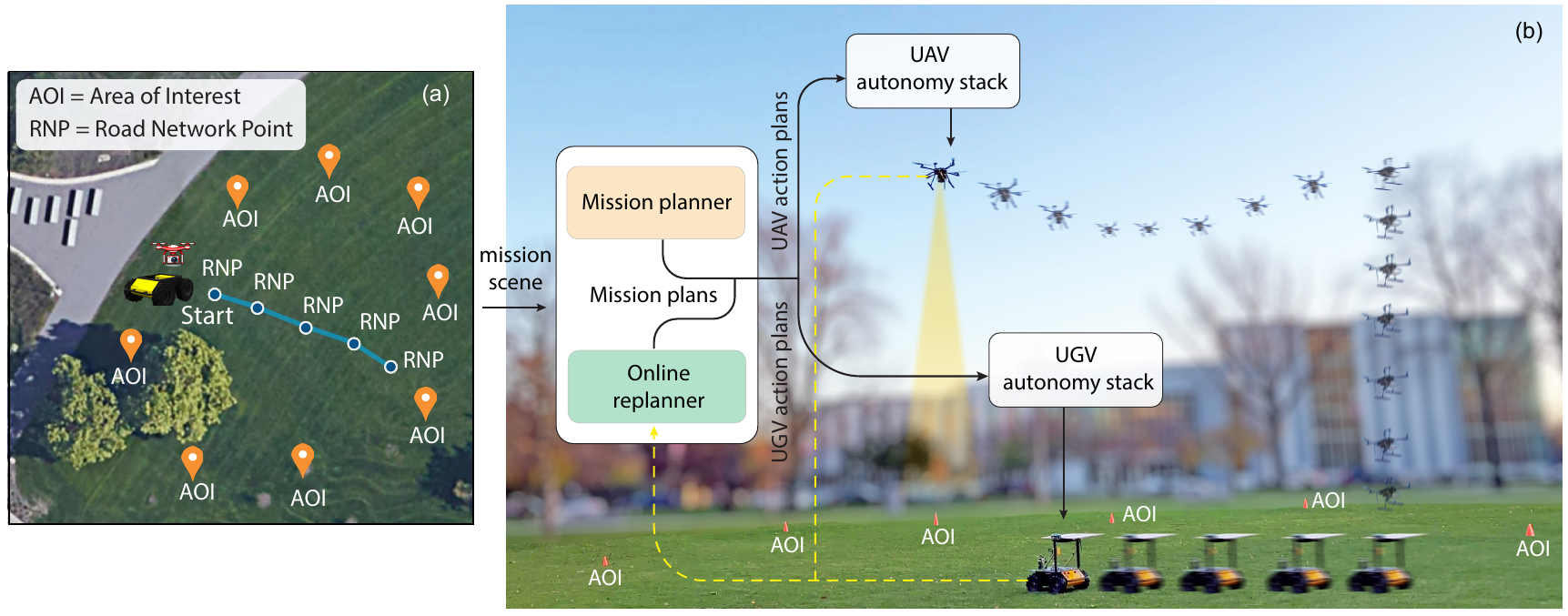}
\caption{Conceptual illustration of the UAV–UGV cooperative mission framework.
The mission planner computes coordinated routes for an endurance-limited UAV and a road-constrained UGV to service a set of distributed Areas of Interest (AOIs). The UAV performs aerial visits, while the UGV travels along the road network to designated rendezvous points (RNP) to support recharging. Onboard autonomy stacks and an online replanner enable dynamic rendezvous, adaptive mission execution, and robust operation under real-world uncertainties.}
\label{fig:concept_figure}
\end{figure*}

Motivated by the operational advantages of heterogeneous aerial–ground systems, this paper investigates an Intelligence, Surveillance, and Reconnaissance (ISR) mission executed by a cooperative UAV–UGV pair. In this scenario, the UAV must visit spatially distributed Areas of Interest (AOIs) (see Fig.~\ref{fig:concept_figure}a), but the mission duration exceeds its flight endurance. The UGV therefore travels along the road network to provide mobile recharging at designated rendezvous points, requiring precise temporal and spatial coordination to prevent battery depletion. Furthermore, real-world hardware execution is susceptible to environmental uncertainties. As illustrated in Fig.~\ref{fig:concept_figure}b, achieving reliable cooperation under these vehicular constraints requires not only sophisticated mission planning but also robust onboard autonomy and online replanning to handle real-world disturbances, including wind effects, sensing noise, and communication latency.

To address these challenges, we present a comprehensive mission-planning framework that solves the cooperative routing problem for an endurance-limited UAV supported by a mobile ground agent. Bridging the gap between theoretical planning and practical deployment, we introduce a complete hardware–software pipeline. This includes a standardized mission API, modular autonomy stacks for both agents, and a lightweight, online rendezvous-aware replanner designed to ensure robust execution on physical platforms. We validate the full system through extensive open-field experiments under varying task configurations, demonstrating the reliability and effectiveness of the proposed framework in realistic outdoor environments.

\section{RELATED WORKS}

Research on aerial--ground cooperation spans multi-agent planning, energy-aware routing, learning-based control, and real-world deployment. This section reviews relevant work and identifies the gap addressed by our integrated planning and autonomy framework.

\subsection{UAV--UGV Collaborative Tasking}

Aerial–ground cooperation has been explored in diverse mission settings, including routing,
refueling, target tracking, and collaborative mapping. Early work by Phan and Liu
\cite{phan2008cooperative} introduced a cooperative wildfire-intervention framework in which a
UAV guides a ground robot via a hierarchical leader–follower control strategy. Tokekar et
al.~\cite{tokekar2016sensor} developed a symbiotic sensing system where a UGV deploys a UAV to
collect measurements in regions occluded from ground view, using an information-theoretic
planner that optimizes sensing while ensuring UAV return for recharging. Li et al. \cite{li2016hybrid} proposed a hybrid UAV–UGV path-planning method that uses UAV-generated ground maps and combines genetic global planning with local rolling optimization for real-time UGV navigation.

Cooperative tracking and mapping have also been explored in structured urban and GNSS-denied environments. Yu et al.~\cite{yu2014cooperative} coordinated UAV--UGV trajectories for target tracking under urban occlusions, while Wang et al.~\cite{wang2024online} studied online path planning for a messenger UAV visiting moving ground vehicles under motion uncertainty. Yue et al.~\cite{yue2022aerial} demonstrated collaborative 3D mapping in GNSS-denied settings using the UGV as a computational anchor, and Christie et al.~\cite{christie2017radiation} used UAV-based semantic segmentation to guide UGV sampling for radiological hazard assessment.

These missions require multi-agent task planning to coordinate heterogeneous capabilities. Manyam et al.~\cite{manyam2019cooperative} and Luo et al.~\cite{luo2017two} formulated aerial--ground routing as a combinatorial optimization problem, with Luo et al. \cite{luo2017two} specifically considering a two-echelon setting where the UGV deploys and recovers the UAV. Since UAV endurance remains a key limitation, Maini et al.~\cite{maini2015cooperation, maini2019cooperative} developed fuel-constrained routing strategies that couple UAV paths with UGV refueling rendezvous, while Yu et al.~\cite{yu2019algorithms} provided algorithmic and experimental validation for UAV routing with mobile recharging stations.

More recently, learning-based methods have enabled scalable coordination. Kool et al.~\cite{kool2018attention} introduced attention-based models for routing problems, and Wu et al.~\cite{wu2021reinforcement} applied similar policies to heterogeneous truck-and-drone delivery coordination.

\subsection{Field Experiments in UAV--UGV Collaborative Operation}

While the algorithmic foundations of aerial–ground cooperation are well-established, experimental validation has primarily focused on specific subsystems such as navigation, mapping, or landing, rather than full-mission autonomy. On the control side, Nigam et al.~\cite{nigam2011control} validated a framework for persistent surveillance using multiple UAVs through coordinated flight tests, optimizing loiter patterns for visual coverage. In contrast, Leahy et al.~\cite{leahy2016persistent} developed a formal planning approach satisfying rich temporal logic specifications under battery constraints, although their demonstrations were restricted to indoor motion capture environments.

Early outdoor deployments, such as those by Michael et al.~\cite{michael2013collaborative}, validated the feasibility of heterogeneous teams by deploying air–ground robots to collaboratively map earthquake-damaged buildings. Subsequent field experiments shifted focus toward physical interaction and relative state estimation. Fankhauser et al.~\cite{fankhauser2016collaborative} demonstrated a system where a UAV generated real-time elevation maps to assist a legged robot in navigating unstructured terrain, relying on relative localization to overcome GNSS-denied conditions. Similarly, Daly et al.~\cite{daly2015coordinated} addressed the control challenges of autonomous recovery, implementing precision landing on moving ground vehicles under communication delays.

Recent experimental efforts have begun to integrate richer onboard perception and higher-level reasoning. Lee et al.~\cite{lee2023aerial} advanced this domain by implementing a real-time aerial mapping system that leverages onboard object detection to dynamically generate traversable paths for UGVs in unstructured outdoor environments. Moving beyond geometric commands, Cladera et al.~\cite{cladera2025air} introduced a framework for language-specified missions, where a UAV–UGV team utilizes Large Language Models (LLMs) to interpret user intent and execute semantic exploration tasks in unknown environments. 

Despite these advances, few experimental frameworks seamlessly integrate robust, risk-aware task planning into a real-time, hardware-deployed autonomy stack. While substantial progress has been made in both algorithmic coordination and isolated field demonstrations, existing approaches often lack a unified architecture that couples mission-level planning with real-world vehicle constraints and closed-loop execution. Consequently, prior works tend to be either theoretically rigorous but practically brittle, or experimentally viable but reliant on static behaviors and human supervision. The development of an end-to-end pipeline spanning optimized planning, platform-aware execution, real-time replanning, and hardware validation remains largely unexplored.

To bridge this gap, we present a fully integrated framework for reliable aerial–ground collaboration. Our specific contributions are: \textbf{First}, a Deep Reinforcement Learning (DRL)-based cooperative routing planner that minimizes mission completion time for a fuel-constrained UAV–UGV team, significantly outperforming established heuristic baselines; \textbf{Second}, a standardized YAML-based mission API that enables seamless translation of planner outputs into executable actions across heterogeneous platforms; \textbf{Third}, comprehensive autonomy stacks for both vehicles, leveraging PX4–MAVSDK for offboard aerial control and ROS~2–Nav2 for ground navigation; \textbf{Fourth}, a lightweight, rendezvous-aware online replanner that ensures safety and feasibility under environmental disturbances and timing uncertainties; and \textbf{Finally}, validation through real-world deployment in a $50 \ \text{m} \times 50 \ \text{m}$ outdoor environment, demonstrating the framework’s robustness in coordinated monitoring tasks as well as a representative search-and-rescue scenario.

The remainder of this paper is organized as follows: Section~\ref{sec3} defines the cooperative UAV–UGV mission problem. Section~\ref{sec4} details the proposed DRL-based mission planner and evaluates its performance. Section~\ref{sec5} describes the full system architecture, including the mission API, autonomy stacks, and the rendezvous-aware replanner. Section~\ref{sec6} presents results from the field experiments, and Section~\ref{sec7} discusses the results, concludes the paper, and outlines directions for future work.


\section{MISSION OVERVIEW}
\label{sec3}

We consider a cooperative mission involving a fuel-limited unmanned aerial vehicle (UAV), denoted by $U_a$, and a ground robot (UGV), denoted by $U_g$, tasked with servicing a set of $n$ Areas of Interest (AOIs),
$M = \{m_0, m_1, \ldots, m_n\}$, distributed across an outdoor environment (shown in Fig. \ref{fig:concept_figure}a). The UGV is constrained to travel along a road-network graph $G$, whose vertices correspond to road-network points (RNPs) and whose edges define feasible ground trajectories.

The two platforms exhibit heterogeneous mobility characteristics. The UAV travels at a relatively high velocity $v_a$ but is limited by its finite flight endurance $F_a$. In contrast, the UGV moves at a slower velocity $v_g$ but is not endurance-limited and can traverse the road network. To remain operational, the UAV must periodically rendezvous with the UGV at any RNP to recharge. Each recharging event requires a fixed service duration $T_R$, after which the UAV resumes its sortie. 

At the start of the mission, the UAV takes off from the UGV at the initial depot. The mission concludes once every AOI has been serviced and the UAV performs a final rendezvous and recharge on the UGV. The objective is to minimize the total mission time while ensuring that all AOIs are serviced at least once. The central challenge lies in jointly planning and executing coordinated UAV and UGV sorties such that:
\begin{itemize}
    \item the overall mission duration is minimized;
    \item the UAV completes all assigned visits without violating its endurance constraints;
    \item the UGV synchronizes with the UAV in both time and location to enable timely rendezvous events.
\end{itemize}
Formally, let $T_{\text{mission}}$ denote the total mission duration, measured from initial UAV takeoff to the final recharge event. Let $\pi_a$ and $\pi_g$ represent the UAV and UGV routing and scheduling policies, respectively. The cooperative routing problem is formulated as:
\begin{equation}
\label{eq:opt_problem}
\begin{aligned}
\textbf{Objective:} \quad \min_{\pi_a,\pi_g} \quad & T_{\mathrm{mission}} \\
\mathrm{s.t.} \quad
& t_{\mathrm{sortie}} \leq F_a,\quad 
  \pi_g \subseteq G, \\
& U_a, U_g \text{ meet at the same RNP and time}, \\
& M \text{ fully serviced}.
\end{aligned}
\end{equation}



\section{MISSION PLANNER}
\label{sec4}

\begin{figure*}[]
\centering
\includegraphics[width=0.9\textwidth]{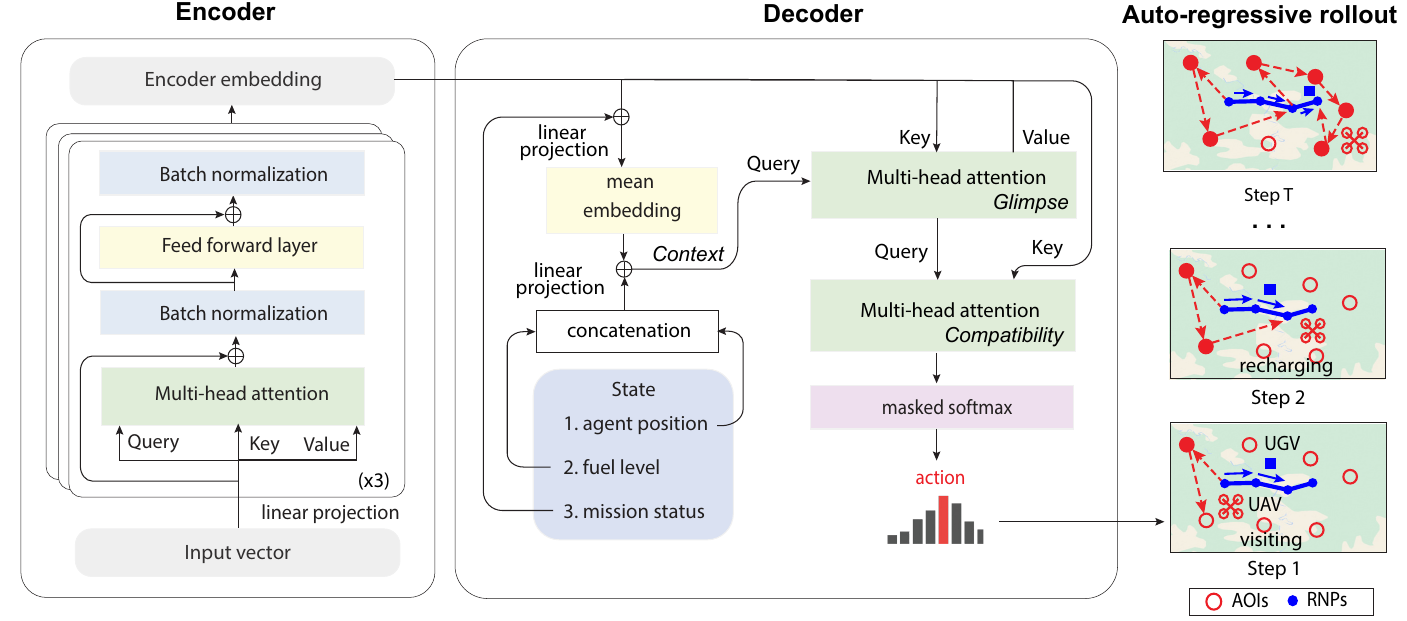}
\caption{Mission planner architecture.
The encoder embeds mission points using multi-head self-attention to capture spatial relationships among AOIs and RNPs. The decoder generates the route auto-regressively by conditioning on the UAV state and attending to the encoder embeddings, while masking infeasible actions at each step. The resulting action sequence specifies successive visitation and recharging decisions during route rollout.}
\label{fig:architecture}
\end{figure*}

The cooperative routing mission requires tightly synchronized planning of UAV sorties, coordinated UGV movements along the road network, and timely UAV--UGV recharging events. To address this, we design a DRL-based mission planner for coordinated decision-making between the UAV and UGV. The planner jointly determines the UAV’s AOI visitation sequence and the road-network point (RNP) where the UAV should rendezvous with the UGV for recharging. The key challenge arises from the UAV’s limited battery capacity and the UGV’s slower, road-constrained motion. The learned policy balances these coupled constraints by reasoning over spatial relationships, energy dynamics, and vehicle motion constraints to minimize overall mission duration.

\subsection{Planner Architecture}

We design a routing policy $\pi_{\theta}$ based on an encoder--decoder Transformer architecture, as shown in Fig.~\ref{fig:architecture}. This structure naturally supports variable-sized mission point sets, captures relational structure among AOIs and RNPs, and enables sequential UAV decision-making. The encoder embeds the spatial configuration of mission points, while the decoder autoregressively constructs a feasible sortie by attending to those embeddings and the evolving mission state.

\subsubsection{Encoder: Learning Mission Point Representations}

Each mission point is represented by a feature vector $o_i=(x_i,y_i,b_i)$, where $(x_i,y_i)$ are normalized coordinates and $b_i\in\{0,1\}$ indicates whether the location is a valid UAV--UGV rendezvous site, i.e., an RNP. The encoder first projects each input vector into a 128-dimensional embedding space.

The projected embeddings are then passed through $L=3$ stacked multi-head self-attention layers, each with $M=8$ heads. At layer $\ell$, attention is computed as:
\begin{equation}
\mathrm{Attention}(Q,K,V)=
\mathrm{softmax}\!\left(\frac{QK^T}{\sqrt{d_k}}\right)V
\end{equation}
where $Q$, $K$, and $V$ denote the query, key, and value matrices, respectively, and $d_k$ is the key dimensionality. This allows each mission point to attend to all others, enabling the encoder to capture spatial dependencies such as AOI clusters, distances to recharge nodes, and the distribution of valid rendezvous locations.

Each attention block is followed by a position-wise feed-forward network with ReLU activation, residual connections, and normalization layers, which improve training stability and gradient propagation. After $L$ layers, each point obtains a contextualized embedding $h_i^L$ that encodes both its local attributes and its relationship to the full mission environment.

\subsubsection{Decoder: Sequential Routing Decisions}

The decoder generates the UAV route one step at a time until all AOIs are serviced and the UAV returns to the UGV for recharge. At decoding step $t$, a context vector is constructed from three components: the embedding of the UAV's current position, the mean embedding of all mission points, and the normalized fuel level $f_t\in[0,1]$. This context serves as the query in a multi-head attention ``glimpse'' over the encoder outputs $\{h_i^L\}$.

The resulting glimpse features are passed through a single-head attention module to compute logits over all candidate next locations. To stabilize training, logits are clipped to $[-C_p,C_p]$ with $C_p=10$. Infeasible nodes, such as previously visited AOIs or recharge nodes that are not needed at the current step, are masked before computing the action distribution:
\begin{equation}
\pi_{\theta}(a_t \mid s_t)=
\mathrm{softmax}\!\left(\mathrm{clip}(h_t)\odot m_t\right)
\end{equation}
where $m_t$ is the feasibility mask.

During training, actions are sampled to encourage exploration. During evaluation, we use both greedy decoding and sampling-based decoding with multiple rollouts to assess solution quality and robustness.

\subsubsection{Training Procedure}

The routing policy is optimized using the REINFORCE policy gradient algorithm, which estimates gradients directly from sampled trajectories. To reduce variance, we use a greedy baseline network $\pi_{\phi}$ that shares the same architecture as the main policy.

For each batch, the policy network $\pi_{\theta}$ samples $B$ trajectories with costs $\{c_1,\dots,c_B\}$, while the baseline $\pi_{\phi}$ generates greedy trajectories with costs $\{c'_1,\dots,c'_B\}$. The policy gradient is computed as:
\begin{equation}
\nabla_\theta J(\theta)=
\mathbb{E}_{\tau\sim\pi_\theta}
\Big[(c-c')\nabla_\theta \log \pi_\theta(\tau)\Big]
\end{equation}

The baseline is updated conservatively using a paired $t$-test on the differences $(c-c')$. The update $\phi\leftarrow\theta$ is applied only when the policy significantly outperforms the baseline, providing a stable and steadily improving variance-reduction mechanism.

This section summarizes the key components of our DRL routing policy. Since the primary contribution of this paper lies in the integration and real-world deployment of UAV--UGV cooperation, we refer readers to our prior work~\cite{mondal2024attention, mondal2025coordinate} for a more exhaustive treatment of the DRL policy design.

\begin{table*}[h]
\tiny
\centering
\caption{Performance of DRL policy on training problem sizes (U15G5, U30G10, U45G15).}
\label{compari}
\renewcommand{\arraystretch}{1.2}
\resizebox{\textwidth}{!}{%
\begin{tabular}{c|ccc|ccc|ccc}
\hline
\multirow{2}{*}{Model} 
& \multicolumn{3}{c|}{U15G5} 
& \multicolumn{3}{c|}{U30G10} 
& \multicolumn{3}{c}{U45G15} \\

& Obj (min) & Gap (\%) & Runtime (sec)
& Obj (min) & Gap (\%) & Runtime (sec)
& Obj (min) & Gap (\%) & Runtime (sec) \\ \hline

DRL (greedy)   & 42.0 & 8.6  & 0.23 & 64.4 & 6.5 & 0.28 & 78.3 & 6.8 & 0.33 \\
DRL (1024)     & 39.0 & 0.8  & 0.24 & 61.9 & 2.3 & 0.46 & 74.6 & 1.7 & 0.56 \\
DRL (10240)    & 38.7 & 0.0  & 2.44 & 60.5 & 0.1 & 4.60 & 73.3 & 0.0 & 6.38 \\
Heuristic \cite{maini2019cooperative}      & 43.9 & 13.5 & 0.01 & 70.4 & 16.3 & 0.01 & 83.5 & 13.9 & 0.08 \\ \hline

\end{tabular}%
}
\end{table*}

\subsection{Model Configuration}

We train the proposed routing policy on three problem sizes: U15G5, U30G10, and U45G15, corresponding to 15/30/45 AOIs and 5/10/15 RNPs, respectively. Each configuration is trained on 5{,}120{,}000 instances generated on-the-fly, using batches of 256 over 100 epochs. We use the Adam optimizer with a learning rate of $10^{-4}$ and an exponential decay factor of 0.995 per epoch. All experiments are conducted on an NVIDIA RTX~4090~Ti GPU. Since cooperative UAV--UGV routing is NP-hard, exact optimization becomes intractable for large missions. Therefore, many heterogeneous routing methods rely on multi-echelon heuristics, such as ``UGV first, UAV second'' or ``truck first, drone second''~\cite{maini2015cooperation, maini2019cooperative, ropero2019terra}. For comparison, we adopt the hierarchical heuristic pipeline of Maini et~al.~\cite{maini2019cooperative} as our baseline. We employ two decoding strategies in our DRL framework: greedy decoding, which selects the maximum-probability action at each step, and sampling decoding, which generates $\mathbb{N}$ trajectories and selects the best solution. In evaluation, we set $\mathbb{N}$=1024 (DRL(1024)) and $\mathbb{N}$=10240 (DRL(10240)).

\subsection{Performance Across Training Problem Sizes}

Table~\ref{compari} summarizes performance across the three training scales.  
Across all settings, the DRL policy substantially outperforms the heuristic baseline in minimizing mission time. The sampling-based variants, particularly DRL(1024) and DRL(10240), achieve the strongest results, reducing the heuristic optimality gap (13--16\%) to below 2.3\%. DRL(10240) consistently attains a zero optimality gap but incurs the highest computational cost, while DRL(1024) offers a better trade-off, producing near-optimal solutions (0.8--2.3\% gap) and running over an order of magnitude faster than DRL(10240). Although DRL(greedy) provides the lowest runtime, it exhibits higher suboptimality (6--9\%).

\subsection{Generalization to Larger Scenarios}

Table ~\ref{gene} evaluates the generalization capability of the learned DRL policy on larger, unseen mission sizes: U60G20 and U75G25. The policy generalizes well beyond its training distribution, consistently outperforming the heuristic baseline by 10--13\% in mission time. DRL(10240) achieves the best objective values with near-zero optimality gap, while DRL(1024) maintains strong performance, achieving gaps below 2.1\% and offering a favorable runtime profile. Although DRL(greedy) provides the fastest inference, its solution quality degrades, yielding gaps in the 10--14\% range.

\begin{table}[h]
\Huge
\centering
\caption{Performance of DRL policy on larger problem sizes (U60G20 and U75G25).}
\label{gene}
\renewcommand{\arraystretch}{1.1}
\resizebox{\columnwidth}{!}{%
\begin{tabular}{c|ccc|ccc}
\hline
\multirow{2}{*}{Model} 
& \multicolumn{3}{c|}{U60G20} 
& \multicolumn{3}{c}{U75G25} \\ 
& Obj (min) & Gap (\%) & Runtime (sec) 
& Obj (min) & Gap (\%) & Runtime (sec) \\ \hline

DRL (greedy)   & 99.3 & 10.5 & 0.378  & 112.7 & 14.0 & 0.41 \\
DRL (1024)     & 91.5 & 1.9  & 0.83   & 100.8 & 2.1  & 1.04 \\
DRL (10240)    & 89.8 & 0.0  & 8.12   & 98.8  & 0.0  & 10.10 \\
Heuristic \cite{maini2019cooperative} & 101.8 & 13.4 & 0.18 & 109.2 & 10.5 & 0.30 \\ \hline

\end{tabular}%
}
\end{table}

Overall, Tables~\ref{compari} and~\ref{gene} demonstrate that the proposed DRL policy, used as the mission planner, consistently outperforms heuristic methods and generalizes robustly to larger-scale cooperative routing tasks, achieving high-quality solutions across all tested scenarios.

\section{INTEGRATED AUTONOMY ARCHITECTURE}
\label{sec5}

The mission planner takes raw scenario inputs, including task-point coordinates and vehicle states, and produces a routing plan that specifies the sequence of visits, as well as the locations and times at which the UAV should rendezvous with the UGV for recharging. To deploy this plan coherently on physical hardware, a coordinated autonomy architecture is required. First, we establish a standardized, mission-specific API representation that ensures the planner receives inputs in a consistent format and produces outputs with a well-defined action structure. Building on this interface, we design distributed autonomy stacks for both agents, enabling the UAV and UGV to reliably translate planned actions into executable controller commands without ambiguity.

\subsection{Mission-Specific API Design}

\begin{figure}[]
\centering
\includegraphics[width=0.5\textwidth]{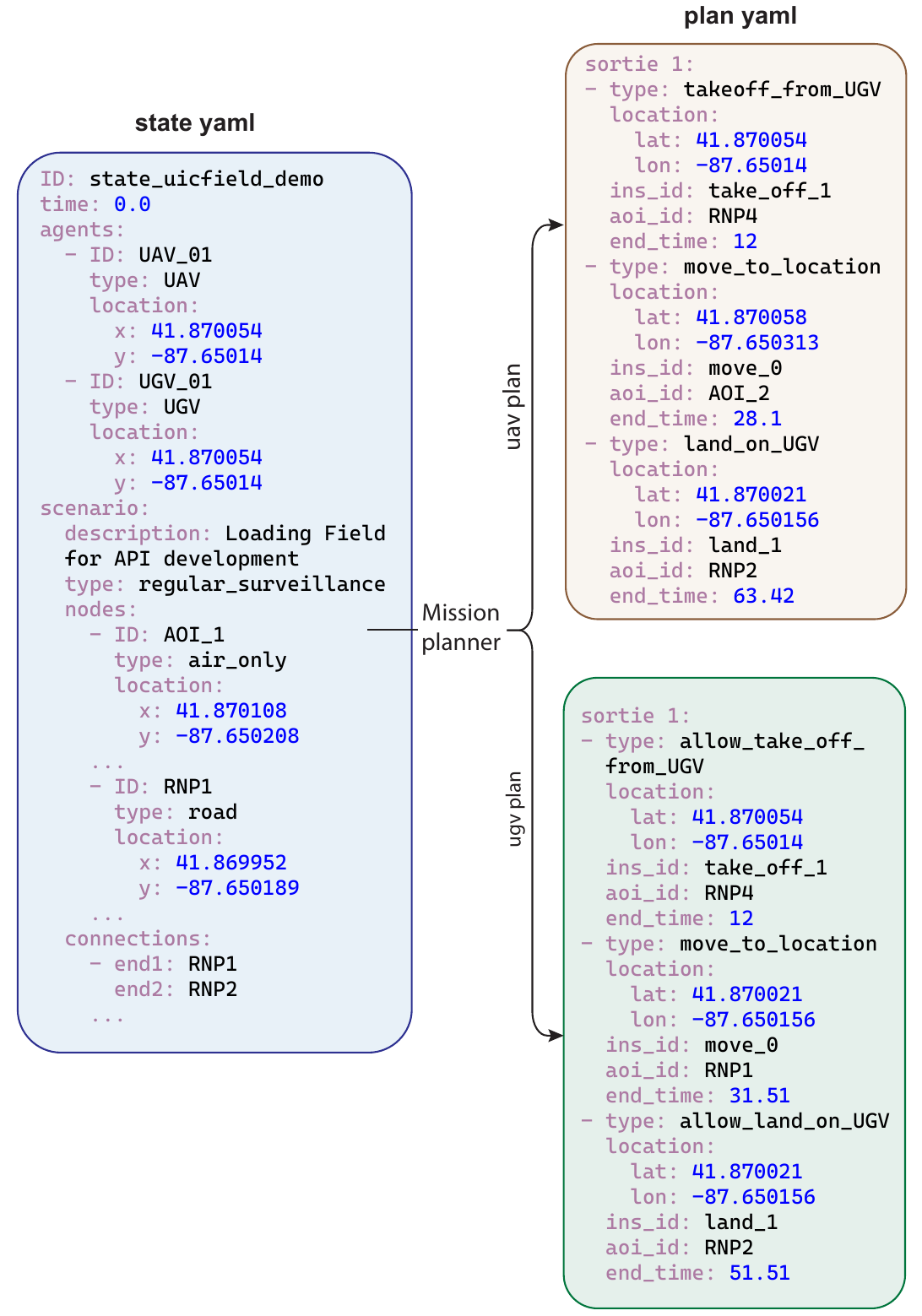}
\caption{Representative snippet of the mission-state YAML provided to the planner and the corresponding UAV and UGV action-plan YAMLs generated for execution.}
\label{fig:api}
\end{figure}

Inspired by prior work~\cite{humann2024data}, we design a task-specific YAML API to standardize how mission information is encoded, interpreted by the planner, and executed on the UAV and UGV platforms. As shown in Fig.~\ref{fig:api}, the API has two complementary layers.

The first layer is the \textit{state-level YAML specification}, which describes the mission environment and initial agent configurations. It includes agent locations, mission metadata, and the set of Areas of Interest (AOIs) and road-network points (RNPs). Each node is defined by a unique identifier, latitude--longitude coordinates, node type, and, when applicable, graph connectivity. This state file provides the planner with a consistent global model before route generation.



\begin{table*}[h]
\centering
\caption{Task-specific YAML API defining the action primitives used by the UAV and UGV during cooperative mission execution.}
\label{task_api}
\resizebox{\textwidth}{!}{%
\begin{tabular}{llll}
\toprule
\textbf{Agent} & \textbf{Action Type} & \textbf{Description} & \textbf{Required Fields} \\
\midrule

\multirow{3}{*}{UAV} 
& \texttt{takeoff\_from\_UGV} 
& UAV performs vertical takeoff from the UGV platform 
& \texttt{location}, \texttt{ins\_id}, \texttt{aoi\_id}, \texttt{end\_time} \\

& \texttt{move\_to\_location} 
& UAV flies to a specified GPS waypoint 
& \texttt{location}, \texttt{ins\_id}, \texttt{aoi\_id}, \texttt{end\_time} \\

& \texttt{land\_on\_UGV} 
& UAV returns and lands on the UGV 
&  \texttt{location}, \texttt{ins\_id}, \texttt{aoi\_id}, \texttt{end\_time} \\
\midrule

\multirow{3}{*}{UGV} 
& \texttt{allow\_take\_off\_from\_UGV} 
& UGV enables UAV takeoff at the specified location 
&  \texttt{location}, \texttt{ins\_id}, \texttt{aoi\_id}, \texttt{end\_time} \\

& \texttt{move\_to\_location} 
& UGV moves to the next waypoint on the road network 
&  \texttt{location}, \texttt{ins\_id}, \texttt{aoi\_id}, \texttt{end\_time} \\

& \texttt{allow\_land\_on\_UGV} 
& UGV enables UAV landing at the platform 
&  \texttt{location}, \texttt{ins\_id}, \texttt{aoi\_id}, \texttt{end\_time} \\
\bottomrule
\end{tabular}
}
\end{table*}

Given this state description, the planner produces an \textit{action-plan YAML} that is lightweight and directly executable by the UAV and UGV control stacks. Each sortie is encoded as an ordered sequence of high-level actions defined by four fields: action \texttt{type}, target \texttt{location}, synchronization identifier \texttt{ins\_id}, and estimated \texttt{end\_time} for temporal coordination. The UAV action plan uses three primitives: \texttt{takeoff\_from\_UGV}, \texttt{move\_to\_location}, and \texttt{land\_on\_UGV}. The UGV uses complementary primitives: \texttt{allow\_takeoff\_from\_UGV}, \texttt{move\_to\_location}, and \texttt{allow\_land\_on\_UGV}. Matched \texttt{ins\_id} tags synchronize takeoff and landing events between the agents.

This minimalist YAML interface contains only the essential information needed for inter-agent communication, sequencing across heterogeneous platforms, and reliable execution of coordinated sorties. Table~\ref{task_api} summarizes the task-level action primitives and their required fields. These primitives are then translated and executed by each agent’s control stack, as described next.

\subsection{Distributed Autonomy Stacks}

\subsubsection{UGV Autonomy Stack}

\begin{figure*}[h]
\centering
\includegraphics[width=1\textwidth]{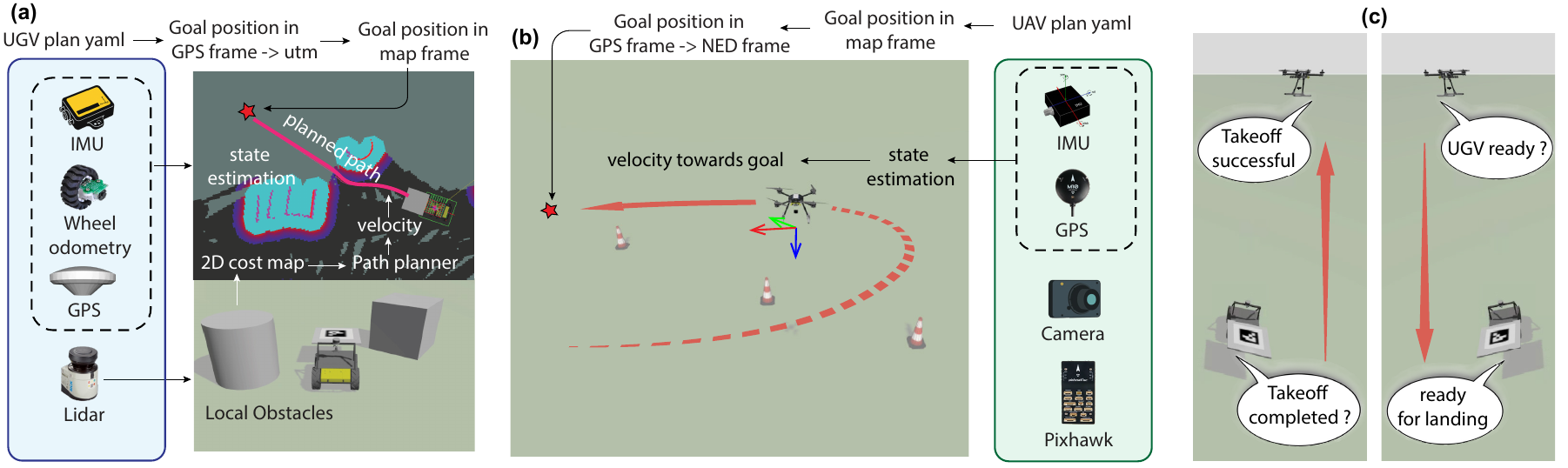}
\caption{Integrated UAV–UGV autonomy and coordination architecture.
(a) UGV autonomy stack based on ROS~2 and Nav2, showing multi-sensor state estimation, costmap construction, and closed-loop navigation for executing planner-generated actions.
(b) UAV closed-loop waypoint guidance using PX4 and MAVSDK, where GPS targets are transformed into the local NED frame to generate offboard velocity commands.
(c) Action-level coordination between the UAV and UGV enabling safe takeoff and landing during rendezvous.}
\label{fig:nav2}
\end{figure*}

The UGV executes its assigned actions through a ROS\,2--based autonomy stack that translates
planner-generated instructions into closed-loop behaviors. Each high-level command in the
action-plan YAML (e.g., \texttt{allow\_takeoff\_from\_UGV}, \texttt{move\_to\_location},
\texttt{allow\_land\_on\_UGV}) triggers a corresponding routine, responsible for motion control
and coordination with the UAV.

For \texttt{allow\_takeoff\_from\_UGV}, the UGV remains stationary and monitors the UAV’s
state via the inter-agent communication channel. Once the UAV has armed, ascended, and cleared
the platform, the UGV marks the action as complete by returning the associated \texttt{ins\_id}.

Robust execution of \texttt{move\_to\_location} relies on accurate state estimation and
autonomous navigation. These capabilities are provided through a multisensor fusion
pipeline and a full Nav2 navigation stack (Fig.~\ref{fig:nav2}a). The UGV employs a dual-EKF
configuration using the \texttt{robot\_localization} package: a \textit{local EKF}, which
fuses high-rate IMU data and wheel odometry to produce a smooth estimate in the
\texttt{odom} frame, and a \textit{global EKF}, which integrates GNSS measurements (converted
to UTM) to anchor the pose in the \texttt{map} frame. The resulting state vector
$(x, y, z, \theta)$ defines the \texttt{map}$\rightarrow$\texttt{base\_link} transform used by
all downstream planning and control modules.

To operate reliably in outdoor environments, the UGV constructs a layered 2D costmap following
Nav2’s modular architecture. A static layer initializes an empty occupancy grid suitable for
unstructured test fields, while a rolling local costmap incorporates real-time lidar scans.
An obstacle layer classifies free and occupied cells, and an inflation layer expands obstacles
according to the UGV footprint and safety margins. This dynamic costmap provides robust
obstacle awareness even in GPS-only deployment scenarios.

For \texttt{move\_to\_location}, the target latitude--longitude waypoint from the action-plan
YAML is converted to UTM coordinates and transformed into the global \texttt{map} frame.
Nav2’s behavior-tree pipeline then executes the motion. The Smac Hybrid-A* global planner \cite{nav2_smac_hybrid_doc}
generates a kinematically feasible path consistent with the UGV’s nonholonomic constraints,
while the MPPI local controller computes velocity commands at 20--30\,Hz by optimizing over
the vehicle dynamics and the local costmap. The UGV follows the resulting trajectory until it
reaches the goal within a 0.3\,m tolerance, after which the behavior tree reports success and
the action executor sends a completion message to the central manager.

During \texttt{allow\_land\_on\_UGV}, the UGV maintains a precise stationary pose and raises a
landing-ready flag once its position is stable. This signal enables the UAV to initiate its
precision landing routine. After each action, the UGV publishes a structured status packet
containing the action identifier, success flag, and completion timestamp. This modular
execution pipeline ensures that the UGV reliably supports rendezvous, recharging operations,
and synchronized multi-robot behaviors throughout the mission.

\subsubsection{UAV Autonomy Stack}

Receiving high-level tasks from the action-plan YAML, the UAV executes its assigned operations through a PX4 and MAVSDK-based action-execution pipeline. This pipeline bridges the symbolic outputs of the planner with continuous control commands, enabling reliable real-world execution. Each YAML-defined instruction (e.g., \texttt{takeoff\_from\_UGV}, \texttt{move\_to\_location}, \texttt{land\_on\_UGV}) is parsed by the UAV’s action wrapper and expanded into a sequence of low-level primitives executed through MAVSDK’s offboard interface.

For \texttt{takeoff\_from\_UGV}, the UAV arms, initiates a PX4-managed takeoff routine, and transitions into a stable hover at a nominal altitude. Once hover stability is confirmed for a fixed dwell period, the UAV reports task completion to the mission coordinator.

During \texttt{move\_to\_location}, the UAV performs closed-loop waypoint guidance (shown in Fig. \ref{fig:nav2}b). The target GPS waypoint is extracted from the action specification, after which the UAV continuously queries its current geodetic state from PX4, converts this to the local NED frame, and computes the relative displacement vector. A proportional velocity controller then generates offboard velocity setpoints streamed to PX4 at 20--50\,Hz to maintain control authority. The motion loop terminates once the geodesic position error remains below a predefined threshold for a sustained dwell time. If the task includes a sensing requirement (e.g., image capture), the UAV captures images upon reaching the waypoint.

For \texttt{land\_on\_UGV}, the UAV first queries the completion flag of the \texttt{allow\_land\_on\_UGV} action from the UGV. Once permission is confirmed, the UAV follows a controlled descent trajectory while estimating its relative pose with respect to the UGV. Depending on the deployment configuration, this relative localization may rely on vision-based fiducials or manual landing cues. Upon touchdown detection, validated through vertical acceleration spikes and near-zero velocity, the UAV disarms and registers the action as complete.

Following each action, the UAV transmits a compact status packet containing the action identifier, success flag, and execution timestamp. This status message enables the central coordinator and the UGV to synchronize subsequent actions and maintain mission-level temporal consistency. The communication protocol governing these status exchanges is described below.

\subsubsection{Communication Stack}

The UAV–UGV team operates over a lightweight bidirectional communication architecture built on UDP and TCP channels. This network links the UAV’s onboard computer (Jetson running MAVSDK with a PX4 interface), the UGV’s onboard computer (Jetson running ROS~2 and Nav2), and a centralized mission manager responsible for executing the mission planner and the rendezvous-aware replanner (described in the next subsection). After generating a mission plan, the central manager broadcasts the action-plan YAML to both agents via UDP, after which each platform executes its assigned actions through its local control stack.

During execution, both agents maintain asynchronous telemetry streams to the central manager at a fixed frequency, sending their low-level vehicle states (e.g., position, nearest waypoint), the currently active action identifier (\texttt{ins\_id}), and task-execution flags. These completion flags are critical for coordinating coupled behaviors such as \texttt{takeoff\_from\_UGV}–\texttt{allow\_takeoff\_from\_UGV} and \texttt{land\_on\_UGV}–\texttt{allow\_land\_on\_UGV} (shown in Fig. \ref{fig:nav2}c). The central manager aggregates these telemetry packets to maintain a global, time-consistent estimate of mission state, which is used to detect deviations from the expected execution timeline.

Replanning is event-driven. Either the UAV or the UGV may issue a replanning request when execution diverges from the nominal schedule, for example, if an action exceeds its expected \texttt{end\_time} or if an agent fails to reach a target waypoint. Upon receiving such a trigger, the central manager initiates a fast replanning cycle using the rendezvous-aware replanner and distributes updated mission plans to both platforms.

This communication loop supports decentralized execution with centralized coordination, ensuring temporal alignment between agents, maintaining global situational awareness, and enabling tightly synchronized rendezvous operations under real outdoor conditions.


\subsection{Rendezvous-Aware Replanner}

\begin{figure*}[]
\centering
\includegraphics[width=0.9\textwidth]{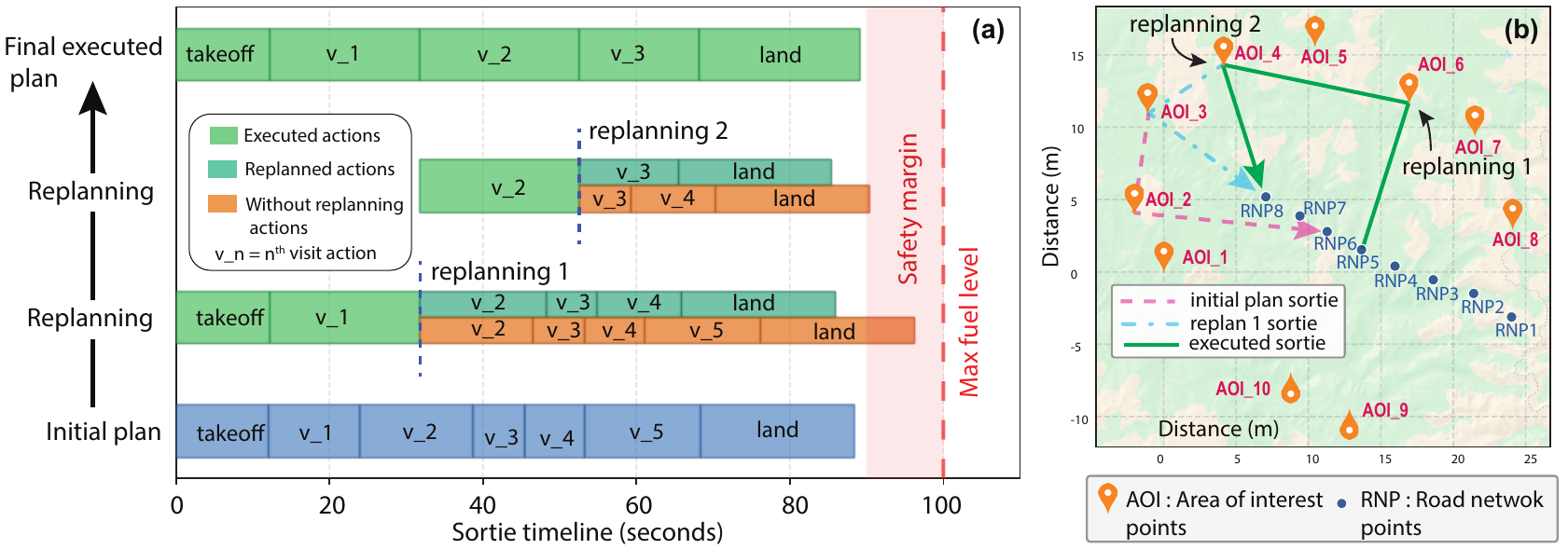}
\caption{Illustration of the rendezvous-aware replanning (RARP) mechanism.
(a) Timeline showing the initial sortie plan, replanned segments, and the final executed sortie, where delayed actions (e.g., \texttt{v\_n}, denoting the $n$th \texttt{move\_to\_location} action) trigger replanning and action trimming to prevent violation of the UAV endurance safety margin.
(b) Spatial view of the corresponding UAV and UGV trajectories, illustrating how trimmed actions and updated rendezvous points (e.g., \texttt{RNP8}) enable a feasible and safe rendezvous despite execution-time delays.}

\label{fig:rarp}
\end{figure*}

\begin{algorithm}[]
\footnotesize
\caption{Rendezvous-Aware Replanning}
\label{alg:replan}

\KwIn{
UAV state $s^{\mathrm{uav}}$ with position $p^{\mathrm{uav}}$; \\
UGV state $s^{\mathrm{ugv}}$ with position $p^{\mathrm{ugv}}$; \\
Current plans $\mathcal{A}^{\mathrm{uav}}, \mathcal{A}^{\mathrm{ugv}}$ with rendezvous $p_{\mathrm{rv}}$; \\
Road graph $\mathcal{G}$; elapsed time $T^{\mathrm{ela}}$; endurance $T^{\mathrm{full}}$; \\
Safety margin $\alpha$ (default $0.90$).
}
\KwOut{Updated plans or an emergency action.}

\textbf{Remaining endurance:}
\[
T_{\mathrm{rem}} = \alpha T^{\mathrm{full}} - T^{\mathrm{ela}}.
\]

\textbf{Phase 1: Check current rendezvous}
\[
T_{\mathrm{uav}}^{\mathrm{plan}} = T_{\mathrm{uav}}(\mathcal{A}^{\mathrm{uav}}),\quad
T_{\mathrm{ugv}}^{\mathrm{rv}} = T_{\mathrm{ugv}}(p^{\mathrm{ugv}}\!\rightarrow p_{\mathrm{rv}}).
\]

\If{$\max(T_{\mathrm{uav}}^{\mathrm{plan}}, T_{\mathrm{ugv}}^{\mathrm{rv}}) \le T_{\mathrm{rem}}$}{
    \Return{$\mathcal{A}^{\mathrm{uav}}, \mathcal{A}^{\mathrm{ugv}}$}
}

\textbf{Phase 2: Rollback search (construct feasible rendezvous)}

\textit{Step 1: Feasible rendezvous set}
\[
\mathcal{V}_{\mathrm{feas}}=\{v\in\mathcal{G} :
    T_{\mathrm{uav}}(p^{\mathrm{uav}}\!\rightarrow v)\le T_{\mathrm{rem}},
    ~T_{\mathrm{ugv}}(p^{\mathrm{ugv}}\!\rightarrow v)\le T_{\mathrm{rem}}\}.
\]

\If{$\mathcal{V}_{\mathrm{feas}} = \emptyset$}{
    go to Phase~3
}

\textit{Step 2: Rollback prefix and pick feasible rendezvous}

\For{$i = |\mathcal{A}^{\text{uav}}|$ \textbf{down to} $1$}{ Let prefix $\mathcal{A}_{1:i}^{\text{uav}}$ end at $v_{\mathrm{end}}$. \\ $T_{\mathrm{pref}} = T_{\mathrm{uav}}(\mathcal{A}_{1:i}^{\text{uav}})$, \; $T_{\mathrm{rem}}' = T_{\mathrm{rem}} - T_{\mathrm{pref}}$. \\[1mm] Candidate rendezvous: \[ \mathcal{V}_{\mathrm{cand}} = \{ v\in\mathcal{V}_{\mathrm{feas}} : T_{\mathrm{uav}}(v_{\mathrm{end}}\!\rightarrow v) \le T_{\mathrm{rem}}' \}. \] \If{$\mathcal{V}_{\mathrm{cand}}\neq\emptyset$}{ $v^* \leftarrow \arg\min_{v \in \mathcal{V}_{\mathrm{cand}}} T_{\mathrm{ugv}}(p^{\mathrm{ugv}} \rightarrow v)$;\\ $\Tilde{\mathcal{A}}^{\text{uav}} = \mathcal{A}_{1:i}^{\text{uav}} \oplus \mathrm{Path}(v_{\mathrm{end}}\!\rightarrow v^*)$.\\ $\Tilde{\mathcal{A}}^{\text{ugv}} = \mathrm{ShortestPath}_{\mathcal{G}}(p^{\text{ugv}}\!\rightarrow v^*)$.\\ \Return{$\Tilde{\mathcal{A}}^{\text{uav}},\,\Tilde{\mathcal{A}}^{\text{ugv}}$} } }

\textbf{Phase 3: Relax safety margin (final attempt)}
\[
\alpha \leftarrow 0.95,\qquad
T_{\mathrm{rem}} = \alpha T^{\mathrm{full}} - T^{\mathrm{ela}}.
\]
Recompute $\mathcal{V}_{\mathrm{feas}}$ and repeat Step~2 of Phase~2.

\textbf{Phase 4: Emergency landing}

$\Tilde{\mathcal{A}}^{\mathrm{uav}} = \mathrm{Land}(p^{\mathrm{uav}})$;\\
$\Tilde{\mathcal{A}}^{\mathrm{ugv}} = 
  \mathrm{ShortestPath}_{\mathcal{G}}(p^{\mathrm{ugv}}\!\rightarrow p^{\mathrm{uav}})$;\\
\Return{$\Tilde{\mathcal{A}}^{\mathrm{uav}},\,\Tilde{\mathcal{A}}^{\mathrm{ugv}}$}

\end{algorithm}

Even when both agents follow the planned action sequence, stochastic environmental factors,
such as wind disturbances, sensing noise, and communication delays, can introduce deviations
from the expected execution timeline. These deviations are critical because the UAV operates
under strict endurance limits, and even small timing mismatches can jeopardize successful
rendezvous with the UGV. To mitigate this risk, we introduce a lightweight Rendezvous-Aware
Replanner (RARP) that runs online and verifies, at each trigger, whether the UAV can still
reach a feasible recharging point. When necessary, the replanner generates a revised pair of
UAV–UGV plans to restore mission feasibility.

The replanner is activated whenever observed execution deviates from the expected timeline
(e.g., an action exceeds its predicted \texttt{end\_time}). RARP then performs a structured
sequence of feasibility checks (Algorithm~\ref{alg:replan}). First, it evaluates whether the UAV’s
remaining endurance is sufficient to complete its assigned actions and still rendezvous with
the originally planned UGV node, assuming the UGV continues along its current route (lines 1–5).
If this condition fails, RARP constructs a set of candidate rendezvous nodes along the road
network that both agents can reach within the UAV’s remaining endurance budget.

A rollback search is then initiated: the replanner progressively truncates the UAV’s future
action sequence and checks whether any shortened prefix can be connected to a feasible
rendezvous point (lines 6–22). If a valid connection is found, synchronized UAV–UGV action
plans are generated and immediately dispatched to both agents. If no feasible connection
exists, the replanner relaxes its endurance safety margin once and repeats the rollback search,
attempting a minimally more permissive rendezvous solution (lines 23–24).

If all attempts fail, the replanner triggers an emergency-landing (lines
24–27). This multi-stage logic provides a fast, robust safety layer that preserves system
integrity under real-world uncertainty and enables resilient, time-sensitive cooperation between
the UAV and UGV.

\subsubsection{Replanner Efficiency}

The performance of the proposed replanning algorithm is evaluated through extensive
experiments using a Clearpath Husky UGV and a PX4 SITL UAV deployed in a Gazebo simulation
environment. Three mission scenarios are constructed, each containing 10 Areas of Interest
(AOIs) and 8 road-network points (RNPs). The UAV is constrained by a 100 \, sec maximum flight
endurance, and a safety factor of $\alpha = 0.9$ is applied, yielding an effective flying
limit of 90 \,seconds per sortie for servicing AOIs and reaching a RNP for recharging.

To assess the robustness of the replanning module, we introduce environmental perturbations
including wind gusts, GPS noise, and a deliberate 15\% reduction in the UAV’s nominal cruise speed relative to the value assumed during planning. Performance is evaluated using two metrics. A sortie is labeled \emph{risky} when its planned duration exceeds 80 \,seconds, placing it near the endurance boundary. The \textit{energy margin violation rate} measures the fraction
of risky sorties whose executed duration exceeds the 90 \,seconds effective limit. The
\textit{deviation rate} quantifies the percentage difference between planned and executed durations for these risky sorties.

As summarized in Table~\ref{tab:replan_results}, the replanning framework significantly
improves sortie reliability under disturbance. The \textit{energy margin violation rate} decreases
from 83.33\% (without replanning) to 20.00\% (with replanning), while the average \textit{deviation
rate} drops by an order of magnitude, from 14.3\% to 1.43\%. These results demonstrate that
the rendezvous-aware replanner effectively mitigates timing drift and keeps sorties within
endurance bounds even under challenging operating conditions.

\begin{table}[h]
\caption{Evaluation of the rendezvous-aware replanning (RARP) module.}
\label{tab:replan_results}
\renewcommand{\arraystretch}{1.4}
\Huge
\resizebox{\columnwidth}{!}{%
\begin{tabular}{ccc}
\hline
 & \textbf{Without replanning} & \textbf{With replanning} \\ \hline
Avg. \textit{energy margin violation rate} (\%) & 83.33 & 20.00 \\
Avg. \textit{deviation rate} (\%) & 14.3 & 1.43 \\ \hline
\end{tabular}%
}
\end{table}

Figure~\ref{fig:rarp} illustrates the behavior of the replanner in a representative example.
In this scenario, the planned sortie requires 88 \ seconds of flight time, which is close to the effective 90 \ seconds budget. Consequently, any delay pushing an action past
its planned \texttt{end\_time} triggers replanning.

As shown in Fig.~\ref{fig:rarp}a, the first deviation occurs after action \texttt{v\_1}
(the initial \texttt{move\_to\_location}). Because execution lags behind schedule, the
replanner is activated. The original rendezvous point, \texttt{RNP6}, becomes infeasible due
to the UAV’s remaining endurance, so the replanner trims action \texttt{v\_5} and selects a
feasible alternative rendezvous point at \texttt{RNP8}. Updated synchronized plans for both
agents are then generated.

A second replanning event occurs after action \texttt{v\_2}, where execution again deviates
from the expected timeline. In this case, trimming action \texttt{v\_4} restores feasibility,
allowing the rendezvous to remain at \texttt{RNP8}. These adjustments are visible in both the
timeline representation (Fig.~\ref{fig:rarp}a) and the spatial sortie trajectories
(Fig.~\ref{fig:rarp}b).

Overall, Fig.~\ref{fig:rarp} highlights how RARP prevents sorties from entering the
safety-margin region or exceeding the UAV’s endurance constraint.

\section{Field Deployment and Experimental Results}
\label{sec6}

\begin{figure}[!b]
\centering
\includegraphics[width=0.35\textwidth]{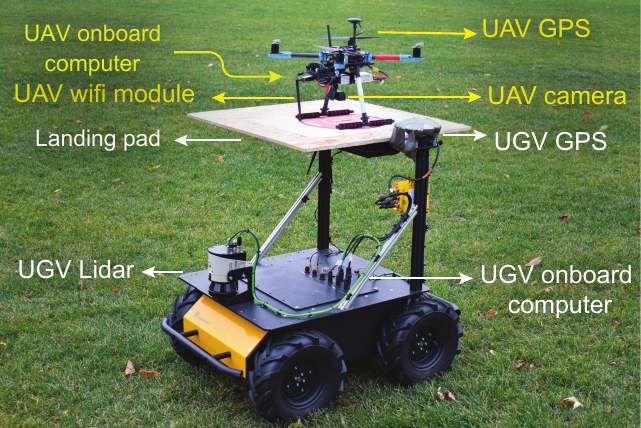}
\caption{Hardware platforms used in the field deployment.
The UAV executes PX4- and MAVSDK-based offboard control, while the UGV operates a ROS~2-based autonomy stack.}
\label{fig:uav-ugv-system}
\end{figure}

We deploy the integrated mission-planning pipeline and autonomy architecture, including the mission planner, standardized mission API, the rendezvous-aware replanner, and the UAV–UGV autonomy stacks, in a real-world UAV–UGV cooperative routing task using a 1~UAV–1~UGV system (shown in Fig. ~\ref{fig:uav-ugv-system}) .

\subsection{Hardware Setup}

\textbf{UAV Platform:}
The aerial platform is a custom-built X500 quadrotor equipped with a Pixhawk~6C flight controller running the PX4 autopilot stack. Our designed onboard autonomy control stack and perception modules are hosted on an NVIDIA Jetson Nano. The sensor suite includes an M10 GNSS module, a downward-facing RGB camera for data collection and visual landing assistance, and an IMU. Communication is facilitated through a long-range telemetry radio and a high-throughput Wi-Fi module for inter-agent coordination.

\textbf{UGV Platform:}
The ground agent is a Clearpath Husky UGV, a four-wheel-drive, four-wheel-steering mobile platform designed for reliable outdoor operation. Computation for our autonomy control is distributed between an NVIDIA Jetson Nano (GPU) and an auxiliary CPU dedicated to low-level control and sensor processing. Perception and state estimation rely on a high-precision SwiftNav GNSS module, an IMU, and a SICK~LMS 2D LiDAR for obstacle avoidance. The software stack is built on ROS~2, utilizing Nav2 for autonomous navigation and for handling UGV–UAV coordination. To facilitate aerial recovery, a custom $30 \ \text{inch} \times 30 \ \text{inch}$ landing pad is mounted on the vehicle’s top deck to support landing and simulated recharging.

\textbf{Communication Network:} 
Inter-agent and ground control communication is established via a dedicated 5~GHz Wi-Fi
network connecting both robots to a central mission computer (Ubuntu~22.04). This link
maintains a telemetry update rate of 30--50~Hz, supporting high-frequency state
synchronization and real-time replanning.

\begin{figure*}[]
\centering
\includegraphics[width=0.9\textwidth]{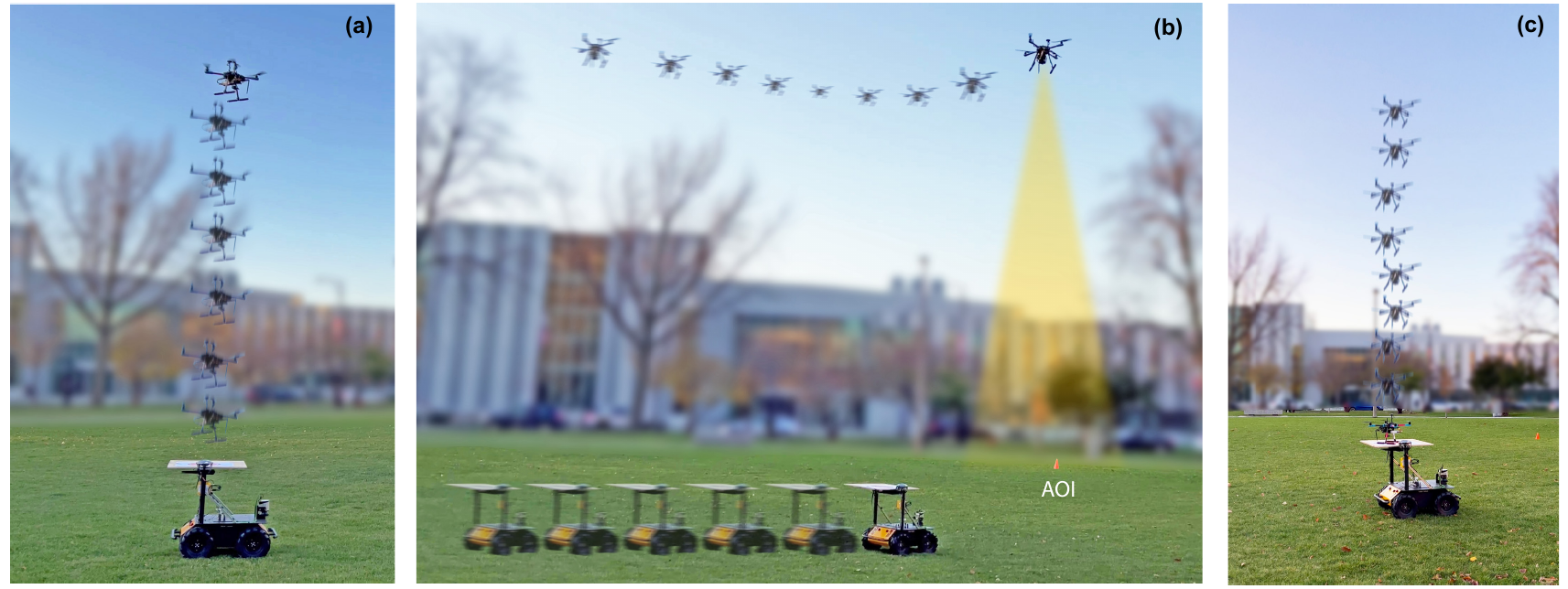}
\caption{Sequence of the field experiment demonstrating the three core tasks of the cooperative mission. The composite images illustrate: (a) The UAV executing an autonomous takeoff from the UGV platform; (b) Synchronized aerial–survey and ground–relocation, where the UAV visits an AOI, while the UGV advances to the next RNP for rendezvous; and (c) Precision landing on the UGV deck for simulated recharging. (video: \href{http://tiny.cc/4mrw001}{\textcolor{blue}{http://tiny.cc/4mrw001}})}
\label{fig:hardware_exp}
\end{figure*}

\begin{figure*}[]
\centering
\includegraphics[width=0.9\textwidth]{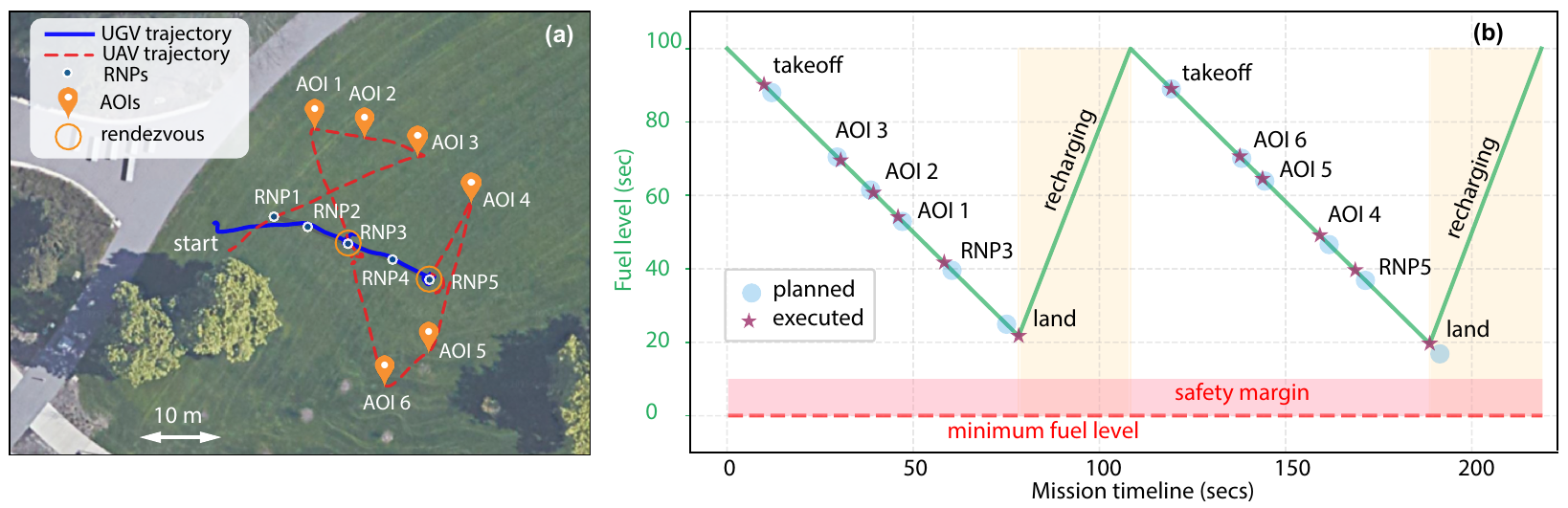}
\caption{Qualitative analysis of the experimental field trial. (a) Trajectories of the UGV and UAV, where the UGV follows the road network (solid blue) while the UAV branches off to visit six Areas of Interest (dashed red). (b) The corresponding energy profile (fuel level vs.\ time). The close alignment between the planner's predicted timeline (blue dots) and the executed task sequence (purple stars) demonstrates the system's ability to maintain feasibility and respect safety margins under real-world conditions.}
\label{fig:trajectory_exp}
\end{figure*}

\subsection{Outdoor Environment}

We validate the complete hardware–software pipeline in an outdoor experimental setting to demonstrate end-to-end cooperative autonomy. Both agents execute fully autonomous onboard stacks while continuously exchanging telemetry, mission status, and replanning requests with a central server hosted on a laptop. This laptop runs the mission planner and online replanner, which generate and dispatch task plans to the agents. Importantly, the mission planner operates in an event-triggered receding-horizon manner. At each UAV–UGV rendezvous event, the current mission state is updated using the latest vehicle states and completed task information, and a new cooperative plan is generated over the remaining AOIs. This design enables adaptive mission execution under changing conditions.

The mission profile consists of coordinated sorties. The UAV is autonomously deployed from the UGV, visits assigned Areas of Interest (AOIs) for data collection, and returns to the UGV to simulate battery recharging. In the field experiments, the final precision landing on the small landing pad is performed manually. Vision-based autonomous landing is validated in Gazebo SITL simulation and will be replaced by a more robust onboard precision-landing system in future field iterations. Concurrently, the UGV traverses the road network to reach designated road-network points (RNPs), ensuring synchronized rendezvous for UAV recharging.

The testing ground covered a $50 \ \text{m} \times 50 \ \text{m}$ region configured with six AOIs and five RNPs. The UAV’s energy constraint is modeled with a nominal endurance of 100~ \ sec; applying a safety factor of $\alpha = 0.9$, the effective flight budget is limited to 90~ sec per sortie. Experiments are conducted on flat grassy terrain under moderate wind conditions (12–24~km/h), with the UAV maintaining a nominal cruise altitude of 7~m.

\subsection{Results and Analysis}

Figure~ \ref{fig:hardware_exp} presents a representative sequence of the cooperative mission executed during field trials. At the start of the mission, the central planner generates an optimized action schedule based on the initial scene state on the laptop. These action plans are subsequently dispatched to the respective agents for synchronized execution. As illustrated, the mission proceeds in distinct phases: the UAV departs from the UGV to survey assigned Areas of Interest (AOIs), while the UGV simultaneously navigates the road network to reach the designated Rendezvous Point (RNP). Upon completing its survey tasks, the UAV executes a precision landing on the UGV at the rendezvous location. A simulated recharging phase is then enforced, during which the UAV remains docked for a fixed service period of 30 sec before initiating the next sortie.

Figure~ \ref{fig:trajectory_exp} provides a qualitative analysis of the agent trajectories and resource utilization. The left panel maps the global paths taken by both agents; for the given environment, the system autonomously generates and executes a two-sortie mission profile to cover all six AOIs. The right panel plots the UAV's energy profile (approximated as remaining flight time) throughout the experiment. Two critical observations can be derived from this data: first, the mission is successfully executed without violating the safety fuel threshold (red dashed line); and second, the temporal alignment between the \textit{planned} state transitions (blue dots) and the actual \textit{executed} states (red stars) is highly accurate, demonstrating the reliability of the system under real-world constraints.

\begin{figure}[]
\centering
\includegraphics[width=0.4\textwidth]{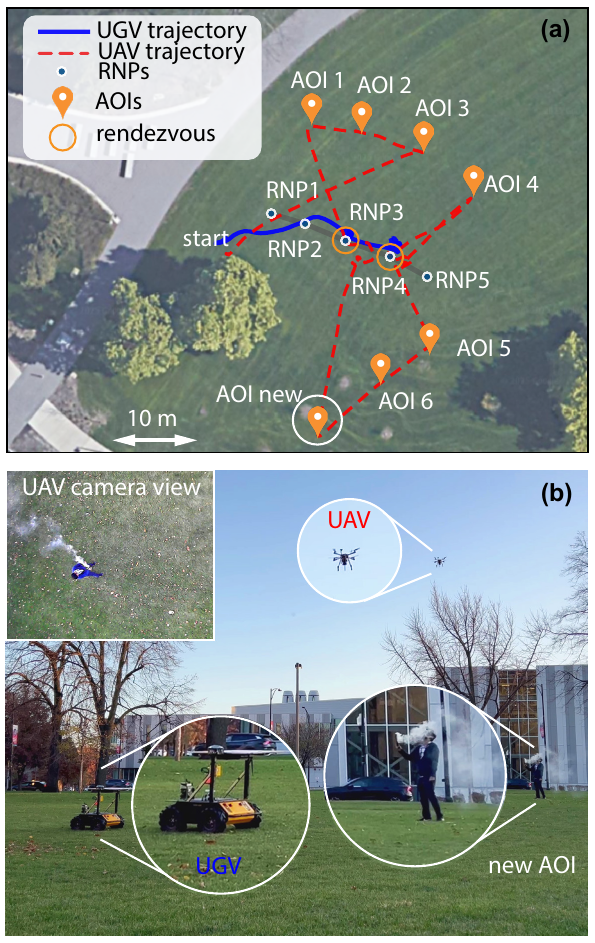}
\caption{Dynamic mission execution with mid-mission task insertion. 
(a) Revised cooperative routing plan after a new AOI is introduced during a UAV--UGV rendezvous. The updated plan adds an additional UAV sortie to service the new AOI while maintaining coordination with the UGV. 
(b) Field deployment corresponding to the updated mission plan, showing the UAV, UGV, and newly introduced AOI. (video: \href{http://tiny.cc/4mrw001}{\textcolor{blue}{http://tiny.cc/4mrw001}})}
\label{fig:dynamic visit}
\end{figure}

\textbf{Dynamic mission execution:}
We also evaluate the system’s ability to handle dynamic task updates in which a new AOI is introduced mid-mission. During a UAV–UGV rendezvous event, the mission planner is re-invoked with the newly added AOI included in the state description. The updated plan incorporates this additional target, and the UAV subsequently visits it during its next flight segment (see
Fig.~ \ref{fig:dynamic visit}), resulting in a modified mission pattern relative to the previous trial. This scenario also requires an additional sortie to ensure that all AOIs are fully serviced.

\subsection{Sample Autonomous Search-and-Rescue Mission}

We validate the proposed planning framework and distributed autonomy architecture through a representative autonomous search-and-rescue (SAR) mission conducted in a field environment. This experiment serves as a proof of concept for deploying the system in post-disaster response scenarios, where rapid situational awareness and coordinated aerial–ground operations are critical.

To emulate a realistic disaster setting, an artificial hazard scene is constructed within the survey area using smoke sources, hazard cones, and human mannequins. During the mission, the system leverages a multimodal vision–language model (VLM) to reason over aerial imagery captured by UAV while surveying the environment and to identify potential hazards relevant to disaster response.

\begin{figure*}[]
\centering
\includegraphics[width=1.0\textwidth]{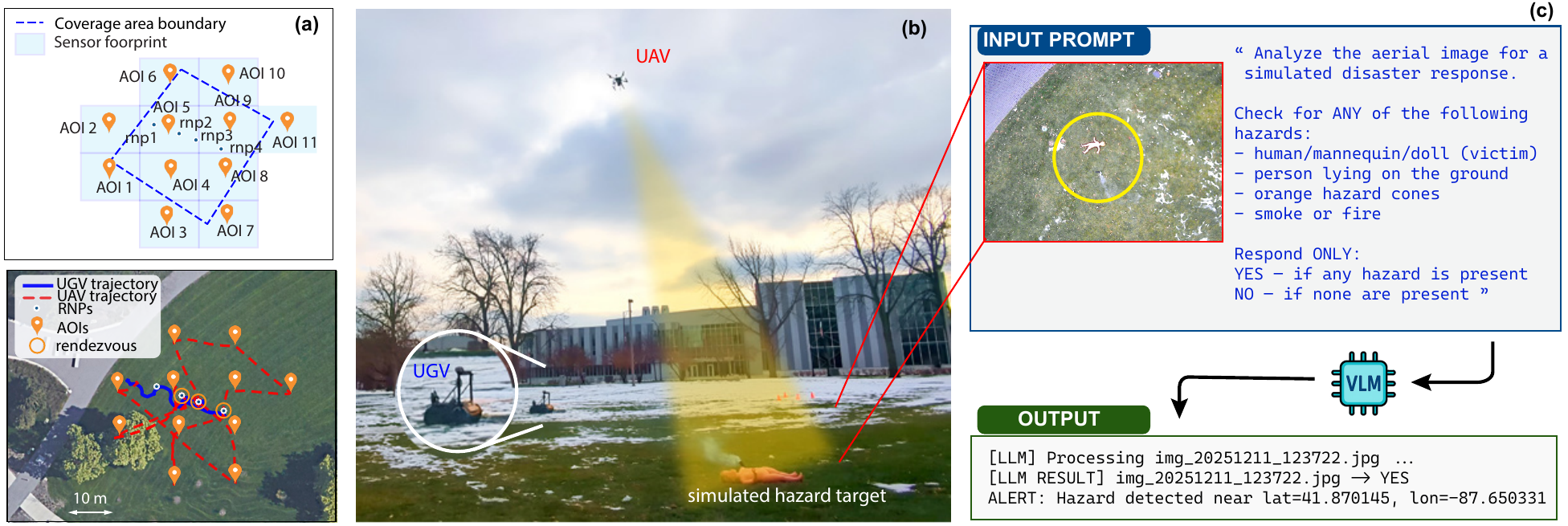}
\caption{Autonomous UAV--UGV search-and-rescue mission with vision--language-based hazard detection.
(a) Coverage planning, where the survey region is discretized using the UAV camera footprint, cell centroids are treated as AOIs, and a cooperative UAV--UGV route is generated.
(b) Field deployment, where the UAV visits each AOI while coordinating rendezvous and recharging with the UGV, and transmits aerial images with geolocation metadata to the mission manager.
(c) Vision--language model pipeline, where a task-specific disaster-response prompt is used to identify potential hazards, such as victims, smoke, or safety markers. Detected hazards are georeferenced using GPS coordinates to support downstream emergency response planning. (video: \href{http://tiny.cc/4mrw001}{\textcolor{blue}{http://tiny.cc/4mrw001}})}
\label{fig:snr}
\end{figure*}

Unlike conventional SAR formulations that rely on predefined Areas of Interest (AOIs), the proposed approach assumes no prior knowledge of hazard locations. Instead, the survey region is discretized based on the ground footprint of the UAV’s onboard camera. The centroids of the resulting discretized cells are treated as AOIs (Fig.~\ref{fig:snr}a), which the UAV visits sequentially to perform systematic coverage and image acquisition.

A coordinated mission plan is generated for the UAV–UGV team by the mission planner, explicitly accounting for the UAV fuel constraint, and rendezvous-based recharging operations. During execution, the UAV captures aerial images at each AOI and transmits them to the central laptop along with associated metadata, including GPS coordinates. These images are analyzed using a vision–language model (VLM), implemented via OpenAI-4o API–based inference \cite{openai2024api}, with a task-specific prompt tailored for post-disaster assessment. This enables the detection of hazards such as victims, smoke, fire, and safety markers. The VLM produces a binary decision indicating the presence or absence of hazardous elements in each scene.

Throughout the mission, the UAV and UGV continuously coordinate to ensure sustained operation under the UAV’s limited flying restriction by synchronizing UAV sorties with UGV-based recharging. Using this framework, the system successfully identifies multiple simulated hazards, including smoke, human mannequins, and hazard cones, within the constructed disaster environment, as shown in Fig.~\ref{fig:snr}b and~\ref{fig:snr}c.

\section{DISCUSSION, CONCLUSION, \& OUTLOOK}
\label{sec7}

\subsection{Discussion}
The results validate the effectiveness of our full-stack approach to aerial--ground collaboration, bridging theoretical optimization and reliable physical deployment. This performance stems from four integrated components: high-quality global planning, standardized interfacing, robust online replanning, and hardware-aware execution.

Reliable coordination begins with an efficient global plan. As shown in Table~\ref{compari}, the proposed DRL policy consistently outperforms multi-echelon heuristics by capturing the coupling between spatial routing and temporal refueling. Table~\ref{gene} further shows that the policy generalizes to larger, unseen scenarios with up to 75 AOIs while maintaining low optimality gaps, demonstrating its scalability for cooperative multi-agent missions.

A key deployment challenge is translating high-level planning decisions into platform-specific control commands. The mission-specific YAML API addresses this by standardizing state representations and action outputs. The autonomy stacks and communication pipeline then connect these high-level actions with low-level execution mechanisms, enabling reliable field deployment.

Because real-world execution deviates from nominal timelines due to wind, sensing noise, and communication delays, online correction is essential. Table~\ref{tab:replan_results} and Fig.~\ref{fig:rarp} show that, without the Rendezvous-Aware Replanner (RARP), sorties exhibit an 83.33\% energy-margin violation rate. With RARP, this drops to 20.00\%, as infeasible actions are trimmed and rendezvous plans are adjusted. This confirms that static global plans alone are insufficient for dependable field operation.

The outdoor experiments provide final validation of the integrated system. Fig.~\ref{fig:trajectory_exp} shows close agreement between predicted and executed fuel usage, supporting the physical fidelity of the planning model. Fig.~\ref{fig:dynamic visit} demonstrates mid-mission AOI insertion, re-optimization, and autonomous execution of the updated plan. Finally, Fig.~\ref{fig:snr} shows deployment in a representative search-and-rescue scenario. Together, these results demonstrate the framework’s readiness for dynamic and realistic field environments.

\subsection{Future Work \& Open Challenges}

While this work establishes a reliable baseline for aerial--ground collaboration, several directions remain for improving scalability and robustness. First, the current framework considers a single UAV--UGV pair. Extending it to heterogeneous multi-UAV--UGV teams will require decentralized coordination, collision-aware trajectory planning, and tighter integration of multi-robot safety mechanisms for real-world deployment. Second, the Rendezvous-Aware Replanner (RARP) currently restores feasibility through reactive tree search and route trimming. Future work will incorporate predictive models of disturbances, such as wind and communication latency, to enable proactive replanning, early sortie termination, and improved timing robustness. Finally, the system relies on high-precision GNSS for localization and rendezvous. Deployment in GPS-denied or contested environments will require vision-based relative localization, visual servoing, and collaborative SLAM to support autonomous docking without global positioning.

\subsection{Conclusion}

This paper presents an integrated framework for reliable, fuel-constrained UAV--UGV collaboration that connects learning-based planning with field deployment. We develop a DRL-based planner that captures the coupling between routing and refueling, outperforms heuristic baselines, and generalizes to larger unseen instances. To enable dependable execution, we introduce a mission-specific API that standardizes state and action representations across heterogeneous platforms. Using this interface, we implement distributed autonomy stacks for the UAV and UGV and integrate a Rendezvous-Aware Replanner (RARP) to correct execution-time deviations caused by disturbances, sensing noise, and communication delays. The results show that this online safety layer is important for maintaining energy feasibility and reliable rendezvous during field operation. Outdoor experiments validate the complete hardware--software pipeline. The system shows close agreement between predicted and executed energy consumption and adapts online to mission changes, including mid-mission task insertion and a search-and-rescue scenario. Overall, this work provides a practical end-to-end foundation for autonomous aerial--ground collaboration in dynamic field environments.

\bibliographystyle{IEEEtran}
\footnotesize
\bibliography{ref}

\vfill\pagebreak

\end{document}